\documentclass[runningheads]{llncs}
\usepackage[T1]{fontenc}
\usepackage{graphicx}
\usepackage{subcaption}

\usepackage{xcolor}

\usepackage{float}

\definecolor{cmu}{HTML}{050203}

\begin{document}
\title{Leveraging LLMs to Assess Tutor Moves in Real-Life Dialogues: A Feasibility Study}
\titlerunning{Leveraging LLMs to Assess Tutor Moves in Real-Life Dialogues}
\author{Danielle R. Thomas\orcidID{0000-0001-8196-3252}\inst{1}
\and Conrad Borchers\orcidID{0000-0003-3437-8979}\inst{1}
\and Jionghao Lin\orcidID{0000-0003-3320-3907}\inst{1}
\and Sanjit Kakarla\orcidID{0009-0007-6508-8647}\inst{1}
\and Shambhavi Bhushan\orcidID{0009-0004-3695-2334}\inst{1}
\and Erin Gatz\orcidID{0000-0002-6880-5740}\inst{1}
\and Shivang Gupta\orcidID{0000-0002-5713-3782}\inst{1}
\and Ralph Abboud\orcidID{0000-0002-2332-0504}\inst{2}
\and Kenneth R. Koedinger\orcidID{0000-0002-5850-4768}\inst{1}
}

\authorrunning{D. R. Thomas et al.}

\institute{
Carnegie Mellon University\\
\email{\{drthomas,cborchers,shivangg,koedinger\}@cmu.edu}\\
\email{\{jionghal,sanjitk,shambhab,egatz\}@andrew.cmu.edu}
\and
Learning Engineering Virtual Institute\\
\email{rabboud@levimath.org}
}
\maketitle 

\begin{abstract}
Tutoring improves student achievement, but identifying and studying what tutoring actions are most associated with student learning at scale based on audio transcriptions is an open research problem. This present study investigates the feasibility and scalability of using generative AI to identify and evaluate specific \textit{tutor moves} in real-life math tutoring. We analyze 50 randomly selected transcripts of college-student remote tutors assisting middle school students in mathematics. Using GPT-4, GPT-4o, GPT-4-turbo, Gemini-1.5-pro, and LearnLM, we assess tutors' application of two tutor skills: delivering effective praise and responding to student math errors. All models reliably detected relevant situations, for example, tutors providing praise to students (94-98\% accuracy) and a student making a math error (82-88\% accuracy) and effectively evaluated the tutors' adherence to tutoring best practices, aligning closely with human judgments (83-89\% and 73-77\%, respectively). We propose a cost-effective prompting strategy and discuss practical implications for using large language models to support scalable assessment in authentic settings. This work further contributes LLM prompts to support reproducibility and research in AI-supported learning.

\end{abstract}

\section{Introduction}

Tutoring is among the most impactful learning methods \cite{nickow2020impressive}. Unfortunately, its impact tends to decrease with scale \cite{kraft2024impacts}. Researchers are striving to identify specific \textit{tutor moves} or actions that are most associated with student learning, which is a difficult task typically done manually. Large language models (LLMs) are increasingly being used to analyze instructional dialogue and identify tutor moves \cite{lin2022good}. A recent randomized controlled trial introduced Tutor CoPilot, a human-AI system that provides expert-like guidance to tutors during live math tutoring, leading to significant student learning gains \cite{wang2024tutor}. However, there is a lack of research benchmarking LLMs' ability to accurately evaluate real-life tutor actions. The current study aims to demonstrate the feasibility of identifying and evaluating two tutor moves using LLMs in real-life tutoring. We demonstrate the use of prompt engineering techniques, including common (e.g., zero- and few-shot prompting) and advanced methods (e.g., rationale forcing, self-consistency) \cite{sahoo2024systematic}. Additionally, we address practical and ethical considerations in using LLMs for assessment, balancing cost and performance while ensuring ethical use in education. We answer the following research question (RQ): \textbf{To what extent is it feasible to use large language models to identify and evaluate specific \textit{tutor moves} in real-life tutor dialogues?}

\section{Related Work}
\textbf{Providing Students with Effective Praise and Responding to Student Errors}. We focus on two common tutoring moves: 1) providing students effective praise and 2) appropriately responding to students who have made math errors. For the former, it is known that process-focused praise (e.g., "I like how you worked hard”) is more effective than person-focused (e.g., “You are smart”) or outcome-based praise (e.g., “Great job, you got 100\%”) \cite{kamins1999person}. Regarding responding to students' errors, expert tutors recommend avoiding direct criticism and suggest encouraging students to identify and correct their own mistakes \cite{wood2012role}. Research suggests tutors should acknowledge student attempts without emphasizing errors, guiding students to self-correct \cite{lepper2002wisdom}. 

\textbf{Large Language Models and Prompt Engineering}. LLMs such as GPT-4, Claude, and LLaMa have recently achieved impressive performance on a wide range of language tasks. This present work uses several GPT models, Gemini-Pro, and LearnLM, a fine-tuned Gemini model optimized for educational tasks. LLMs have attracted significant interest in education due to their potential for scalable and cost-effective reasoning. However, they exhibit critical limitations. LLMs are increasingly valuable in education due to their scalability and reasoning capabilities, but they face key challenges: their black-box nature raises safety concerns, and they often hallucinate \cite{zhang2023sirens}, producing confident but false content. To improve reliability, techniques like prompt engineering, few-shot prompting \cite{brown2020language}, and chain-of-thought reasoning \cite{wei2022chain} are used. Self-consistency methods \cite{wang2022self} further boost accuracy by majority voting over multiple outputs. 

\section{Methods}

\textbf{Corpus Description \& Data Pre-Processing}. The dataset consisted of 50 tutor-student audio transcriptions of college students providing remote tutoring to middle school students. Tutors and students communicated during tutoring sessions via video conferencing software with the audio being recorded and transcribed. Most transcriptions are 1:1 tutor and student sessions. Data pre-processing consisted of choosing transcriptions that were between 2 and 11KB. Through exploratory trial-and-error, transcription file sizes smaller than 2KB often did not contain enough content-related dialogue and transcription files larger than 11KB tended to be too large when running transcriptions through GPT-4. The transcription software did not provide identification of “tutor” and “student” within the transcripts. To mitigate this issue, we fed the text as-is by selecting tasks where the speaker can be determined by the target behaviors. 

\textbf{Human Annotation and Coding}. Two expert researchers annotated classifying each transcript by indicating “yes” or “no” for two questions for both tutoring skills. The first question is considered a filter to determine if the tutoring situation applies, and, if found to be satisfied, the second question is an evaluation of the tutors’ moves. Table 1 displays sample tutor utterances within the corpus for the evaluation question of how well the tutor responds. \noindent\textbf{(Praise) Filter:} Does the tutor praise the student? \noindent\textbf{Evaluation}: If so, does the tutor focus on the student’s effort and acknowledge the learning process? \noindent\textbf{(Errors) Filter}: Did the student make a math error? \noindent\textbf{Evaluation}: If so, does the tutor react by not calling direct attention to the error but rather guiding the student to find their own mistake?

\begin{table}[htp]
\caption{Sample tutor utterances with coding rationale for tutors providing praise to students (A) and reacting to students making errors (B).}
\label{tab:combined-table}
\scriptsize
\renewcommand{\arraystretch}{1.2}
\centering
\begin{tabular}{p{5cm}p{7cm}}
\hline
\multicolumn{2}{l}{\textbf{(A) Praise to Students}} \\
\hline
\textbf{Tutor Response} & \textbf{Rationale} \\
\hline
\textit{You all did great working on adding {[}and{]} subtracting fractions… Great work… keep it up!} & Yes/Correct: This tutor response praises the student for their effort on learning identifying specific actions by the student. “Great work” implies effort over outcomes or ability. \\
\textit{Awesome… Great work… Nice… Love it. Love it… Cool!} & Yes/Correct: Although there is no specific reference to content the student is learning, the tutor states “great work,” implying the praise of effort over ability or outcomes. \\
\textit{Yes, that’s perfect… Perfect. Okay. Doing an awesome job so far.} & No/Incorrect. The tutor praises the student by stating “perfect,’ which implies praising for outcome. Saying “awesome job” also implies outcome over “work” acknowledging effort. \\
\hline
\multicolumn{2}{l}{\textbf{(B) Reaction to Student Math Errors}} \\
\hline
\textbf{Tutor Response} & \textbf{Rationale} \\
\hline
\textit{So you’re super close… How can we take this and apply what we’ve been doing so far?} & Yes/Correct: Although the tutor calls attention to the error, it was already indicated as incorrect on the student’s math software. The tutor then guides the student to self-correct. \\
\textit{Are you sure? Two times three? You got that one wrong… Nope, that’s no, that’s not right.} & No/Incorrect: The tutor directly calls attention to the student’s mistake. \\
\textit{So can you explain to me why you got that one wrong?} & Yes/Correct: Although, the tutor calls attention to the student’s error, the student stated in dialogue prior that they got the problem incorrect. The tutor then guides the student to self-correct. \\
\hline
\end{tabular}
\end{table}

Two researchers coded transcripts for filter and evaluation questions. They agreed that tutor praise and student math errors occurred in 44\% and 50\% of cases, respectively. Agreement was high for filter questions—96\% (Cohen's $\kappa$ = 0.92) for praise and 96\% (Cohen's $\kappa$ = 0.80) for error detection. Evaluation question agreement, which was calculated only when both raters agreed the filter condition was met, was lower: 78\% for praise quality and 72\% for error responses. A transcript was marked "correct" if the tutor provided at least one effective praise or appropriate error response.

\textbf{LLM Prompting and Assessing Performance}. 
All grading prompts are provided in the Digital Appendix.\footnote{https://github.com/conradborchers/real-tutor-moves-ectel25} Prompt engineering followed an iterative trial-and-error process. Initial prompts were developed with a temperature of 0 using GPT-4 and later refined using self-consistency prompting—raising the temperature and aggregating multiple outputs via majority vote. Most prompts requested rationale, except for evaluating tutor responses to math errors, where omitting rationale improved performance. We applied these prompts to GPT-4o and GPT-4turbo, Gemini 1.5 Pro, and LearnLM. Minor adjustments were needed for Gemini-based models. For math error detection, prompts were modified to identify tutors' recognition of errors rather than the errors themselves, since transcripts lacked multimodal data. For praise, the criterion prioritized presence of effective praise over frequency to reduce noise from repeated affirmations.

To evaluate performance, we used a scoring system that considers both the filter and evaluation prompt outcomes per skill. A transcript earns 2 points if the model and human coder agree on both prompts, and 1 point if only the filter prompt matches. If both code the filter as -1, the transcript also receives 2 points, as the evaluation prompt is then skipped by design. With 50 transcripts, a perfect model score totals 100. This system enables comparison across tutor skills and identifies which areas models handle more reliably.

\section{Results \& Discussion}

\textbf{RQ: How effective are LLMs in assessing \textit{tutor moves}?} Our findings indicate that LLMs show strong potential for evaluating real-world tutor responses. As shown in Table 2, model accuracy was high—over 94\% for identifying tutor praise and 82\% for detecting student math errors. Using 95\% bootstrap confidence intervals, we observed that all LLMs performed comparably, with no model significantly outperforming others. Overall, LLMs achieved accuracies well above chance, demonstrating their reliability in assessing key tutor behaviors. Beyond detecting tutor moves, LLMs also performed well in evaluating their quality, with scores of 83–89\% for praise and 73–77\% for error responses. The drop in accuracy compared to detection likely reflects the greater complexity of grading, which requires the interpretation of subtle contextual and linguistic cues. This is especially true for error responses, where effective feedback involves indirect guidance rather than direct correction, contributing to lower accuracy.

\begin{table}[ht]
\centering
\caption{Bootstrap (10,000 samples) accuracy scores and 95\% percentile confidence intervals for LLM performance on tutor response classification, by prompt.}
\label{tab:juxtaposed_praise_error}
\renewcommand{\arraystretch}{1.2}
\begin{tabular}{lcccc}
\hline
\textbf{Model} & \multicolumn{2}{c}{\textbf{Filter Prompt}} & \multicolumn{2}{c}{\textbf{Evaluation Prompt}} \\
\cline{2-3} \cline{4-5}
 & Praise (95\% CI) & Errors (95\% CI) & Praise (95\% CI) & Errors (95\% CI) \\
\hline
GPT-4           & 98, (94, 100) & 84, (74, 94) & 89, (82, 95) & 76, (65, 86) \\
GPT-4o          & 94, (86, 100) & 82, (70, 92) & 85, (77, 92) & 73, (62, 84) \\
GPT-4 Turbo     & 96, (90, 100) & 84, (74, 94) & 87, (79, 94) & 76, (65, 86) \\
Gemini 1.5 Pro  & 94, (86, 100) & 88, (78, 96) & 83, (75, 91) & 76, (66, 85) \\
LearnLM         & 94, (86, 100) & 86, (76, 94) & 85, (77, 92) & 77, (67, 87) \\
At-Chance       & 25, (18, 32)  & 33, (18, 32) & 46, (32, 58) & 57, (32, 57) \\
\hline
\end{tabular}
\end{table}

An example of LLM limitations in praise grading involves Tutor 1: “Yeah. You got it. Awesome. Great job.” and Tutor 2: “Awesome. Perfect…keep up the good work.” According to the evaluation rubric, Tutor 1 uses outcome-based praise—“great job” suggests a task-based result—while Tutor 2’s “keep up the good work” emphasizes effort and process. These subtle differences underscore the evaluation prompt’s sensitivity to nuanced language.

\textbf{Insights into LLM and Prompt Use for Assessment}. Few-shot prompting \cite{brown2020language} improved model accuracy by exposing LLMs to both ideal and suboptimal tutor responses, helping them distinguish nuanced feedback like effort-based versus outcome-based praise. Chain-of-thought prompting \cite{wei2022chain} enhanced performance when reasoning was required—especially in praise detection—but decreased accuracy for error recognition, suggesting the utility of rationale depends on task complexity. Prompt chaining \cite{WuTC22} and self-consistency \cite{wang2022self} allowed us to modularize the evaluation process and stabilize predictions, with majority voting proving effective for binary decisions in filter and evaluation stages.

\section{Limitations, Future Work, and Conclusion}
One limitation of this study is the absence of video, screenshots, and chat messages, which may have led to an incomplete view of the tutoring interactions. Identifying speakers in transcriptions was also challenging, making it difficult to distinguish tutors from students. However, as a feasibility study, our aim was to test whether lightly processed transcripts could support tutor move detection and evaluation using LLMs.

Future work will involve improving prompting methods and increasing more transcriptions with human labeling. Our study demonstrates the promise of LLMs in assessing tutors in real tutoring situations. Analyzing tutoring transcripts, our approach achieves high accuracy in determining whether a tutor implemented a best practice, as well as assessing the quality of their implementations. Notably, our results suggest no single LLM significantly outperformed the others for these particular tasks. Our open-source prompting methods and findings contribute a pathway toward evaluating tutoring at scale. 

\section*{Acknowledgments}
This work was made possible with the support of the Learning Engineering Virtual Institute. The opinions, findings and conclusions expressed in this material are those of the authors.

\bibliographystyle{splncs04}
\bibliography{main}

\end{document}